\documentclass[conference]{IEEEtran}
\IEEEoverridecommandlockouts

\usepackage{cite}
\usepackage{amsmath,amssymb,amsfonts}
\usepackage{algorithmic}
\usepackage{graphicx}
\usepackage{textcomp}
\usepackage{xcolor}
\def\BibTeX{{\rm B\kern-.05em{\sc i\kern-.025em b}\kern-.08em
    T\kern-.1667em\lower.7ex\hbox{E}\kern-.125emX}}
\usepackage{booktabs}
\usepackage{url}
\usepackage{multirow}

\begin{document}

\title{Generalizable Detection of AI Generated Images with Large Models and Fuzzy Decision Tree}

\author{
\IEEEauthorblockN{Fei Wu, Guanghao Ding, Zijian Niu, Zhenrui Wang, Lei Yang, Zhuosheng Zhang, Shilin Wang\IEEEauthorrefmark{1}}
\IEEEauthorblockA{\IEEEauthorrefmark{1}Shanghai Jiao Tong University \\
\textit{\{wu\_fei, Departure, yishuihan, blizzard-wang, yangleisx, zhangzs, wsl\}@sjtu.edu.cn}}
\thanks{This work was supported by the National Natural Science Foundation of China (NSFC) under Grant No.~62572315, and by the Science and Technology Commission of Shanghai Municipality under Grant No.~24JG0500302. (Corresponding author: Shilin Wang.)}
}

\maketitle

\begin{abstract}
The malicious use and widespread dissemination of AI-generated images pose a serious threat to the authenticity of digital content. 
Existing detection methods exploit low-level artifacts left by common manipulation steps within the generation pipeline, but they often lack generalization due to model-specific overfitting. 
Recently, researchers have resorted to Multimodal Large Language Models (MLLMs) for AIGC detection, leveraging their high-level semantic reasoning and broad generalization capabilities. 
While promising, MLLMs lack the fine-grained perceptual sensitivity to subtle generation artifacts, making them inadequate as standalone detectors.
To address this issue, we propose a novel AI-generated image detection framework that synergistically integrates lightweight artifact-aware detectors with MLLMs via a fuzzy decision tree. 
The decision tree treats the outputs of basic detectors as fuzzy membership values, enabling adaptive fusion of complementary cues from semantic and perceptual perspectives. 
Extensive experiments demonstrate that the proposed method achieves state-of-the-art accuracy and strong generalization across diverse generative models.
\end{abstract}


\section{Introduction}
\IEEEPARstart{T}{he} remarkable progress of generative image models, particularly that of Generative Adversarial Networks~\cite{goodfellow2020generative} and Diffusion Models~\cite{ho2020denoising} has substantially advanced the field of image generation. 
However, these advances also give rise to significant risks, notably the fabrication of misinformation and other forms of digital deception~\cite{cnn2024deepfake}.
Consequently, the detection of AI-generated images has become imperative to safeguard the authenticity of digital content.

Numerous AIGC detection methods~\cite{ojha2023towards, zhong2023rich, tan2024rethinking, tan2024frequency, chen2024drct, yan2024sanity, zhong2025beyond, yan2024effort, chen2025dual, yang2025d} have been proposed. 
Early approaches primarily focused on pixel-level fingerprints introduced by GAN-based generation, such as texture distribution biases~\cite{zhong2023rich}, upsampling induced artifacts~\cite{tan2024rethinking} and frequency-domain inconsistencies~\cite{tan2024frequency}.
Recent studies have broadened the scope to diffusion-generated images and exploited invariant forgery traces introduced by shared operations in the generation pipeline, enabling more robust and generalizable detection~\cite{chen2024drct, yan2024sanity, zhong2025beyond, yan2024effort}.
However, such reliance often leads to overfitting to specific forgery patterns present in the training data, inherently constraining generalization to images generated by unseen models.

To handle these challenges, recent research has increasingly shifted toward Multimodal Large Language Models (MLLMs), leveraging their strong semantic reasoning and cross-modal representation ability.
A common practice for adapting MLLMs to downstream tasks such as AI-generated image detection is fine-tuning them on labeled datasets associated with specific generative models. 
However, fine-tuning often leads to overfitting to training-specific forgery patterns. 
An alternative is to employ MLLMs in a zero-shot manner using carefully designed and broadly applicable prompts, which better preserves their generalization. 
Nevertheless, MLLMs often lack the fine-grained perceptual sensitivity required to detect subtle generation artifacts, making it challenging to identify forgeries solely based on high-level semantics.

The limited generalization ability of lightweight artifact-aware detectors and the insufficient fine-grained perceptual sensitivity of large semantic models motivate their integration to combine their complementary strengths for more generalizable detection across diverse generative models.

To this end, we propose a novel AI-generated image detection framework that synergistically combines both types of detectors through a fuzzy decision tree~\cite{zhu2024development}.
By considering the outputs of basic detectors as fuzzy membership, the fuzzy decision tree combines the merits of different categories of detectors and yields an adaptive detection rule in the form of fuzzy logic, which allows further interpretation and optimization with arbitrary off-the-shelf AI-generated image detectors. 

\begin{figure}[!t]
  \centering
  \includegraphics[width=\linewidth]{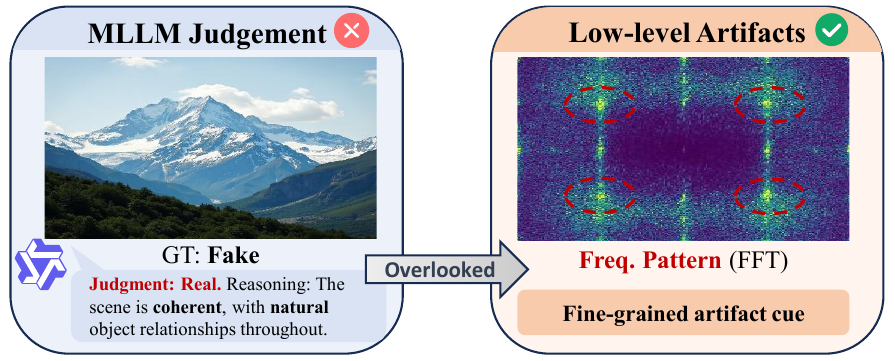}
  \caption{An example where an MLLM fails to detect a fake image due to its insensitivity to fine-grained visual artifacts.}
  \label{fig:case}
\end{figure}

The major contributions of this paper are as follows:

\begin{itemize}

    \item We reveal the overfitting phenomenon of existing detectors, which significantly limits their cross-model generalization performance.

    \item We achieve zero-shot MLLM AI-generated image detection and observe that the detection results from MLLMs are highly complementary to those from lightweight artifact-aware detectors.

    \item We propose a fuzzy decision tree-based AI-generated image detection framework that combines low-level artifact cues and high-level semantic reasoning using fuzzy logic, achieving state-of-the-art performance.
\end{itemize}

\begin{figure*}[!t]
    \centering
    \includegraphics[width=\linewidth]{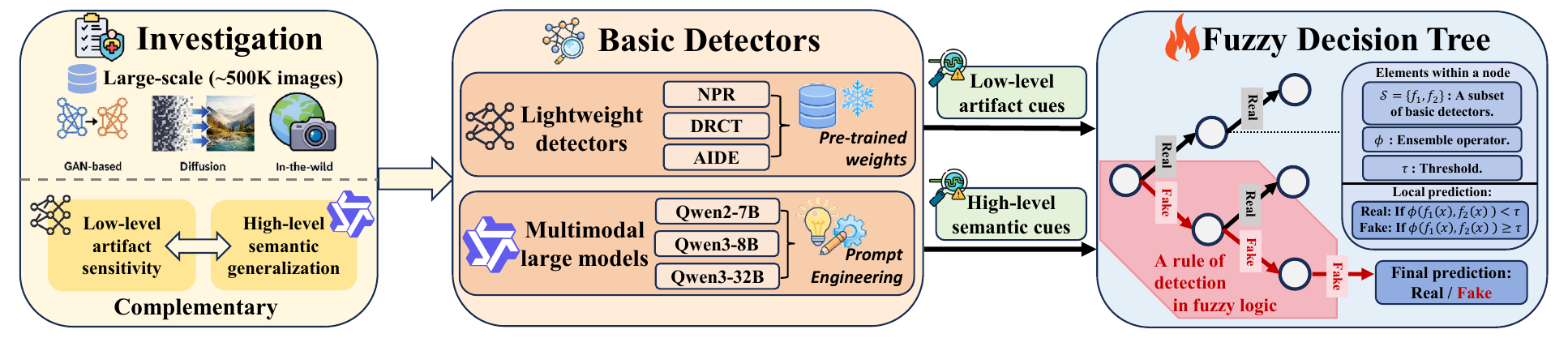}
    \caption{An overview of our proposed method, which integrates lightweight detectors and multimodal large language models via a fuzzy decision tree.
    }
    \label{fig:overview}
    \vspace{-1em}
\end{figure*}
\section{Challenges and Motivations} \label{section:2}

\subsection{Challenges}

Existing AI-generated image detection methods can be divided into two categories: lightweight detectors and multimodal large language models.

Lightweight detectors are efficient and can be categorized into two types: those that capture model-specific low-level artifacts introduced during the generation process~\cite{zhong2023rich, tan2024rethinking}, and those that utilize semantic visual features extracted by pretrained encoders for real/fake classification~\cite{ojha2023towards, liu2024forgery}.

In contrast, MLLM-based approaches leverage the extensive world knowledge and high-level semantic reasoning of large vision-language models. 
FFAA~\cite{huang2024ffaa} and Veritas~\cite{tan2025veritas} mainly address Deepfake detection, whereas FakeShield~\cite{xu2024fakeshield} and SIDA~\cite{HuangHLH00W0C25} focus on image editing forensics. 
Despite these efforts, the application of MLLMs to general-purpose AIGC detection remains underexplored.

Nevertheless, both types of approaches can hardly yield satisfactory performance in practical scenarios due to the following challenges. 

\textbf{Challenge I: Low-level artifact-based detectors lack generalization ability.} 
Images generated by different models contain identifiable low-level forgery traces that vary substantially across generative models. 
As a result, detectors trained on these low-level features are prone to overfitting to the forgery patterns present in the training data, which limits their ability to generalize to new or unseen generative models.

\textbf{Challenge II: MLLMs lack sensitivity to discriminative traces.} 
As illustrated in Fig.~\ref{fig:case}, the strong reliance of MLLMs on global semantic context limits their necessary sensitivity to subtle low-level forgery traces, making it challenging to identify fine-grained artifacts indicative of generated content.

\subsection{Motivations}

Experiments on AIGIBench~\cite{li2025artificial} reveal complementary limitations between lightweight detectors and MLLMs, motivating the integration of their outputs for improved detection performance. 
However, the heterogeneity in output formats limits the effectiveness of vanilla ensemble strategies such as majority voting or logistic regression.

To this end, we introduce a fuzzy decision tree to fuse multiple sources of outputs.
A fuzzy decision tree interprets the heterogeneous outputs of each detector as membership values~\cite{han2023three,zhu2024development}.
Meanwhile, it can exploit the complementary strengths of the sensitivity to fine-grained artifacts of lightweight models and the high-level semantic reasoning ability of MLLMs in an interpretable and logical way. 
\section{Methodology} \label{section:3}
\subsection{Overview}

An overview of our method is shown in Fig.~\ref{fig:overview}.
Concretely, each input is given to several pretrained lightweight detectors and a series of open-source MLLMs. 
Each model provides either a continuous likelihood in $[0,1]$ or a zero-shot binary decision. 
These heterogeneous outputs are passed to a fuzzy decision tree, which is trained with a small collection of labeled samples. 

During training, at each node of the fuzzy decision tree, the optimal fuzzy predicate is selected based on the predictions of all candidate detectors. 
Recursively, the fuzzy decision tree forms a collection of fuzzy logic rules that combines the predictions from all detectors. 

\subsection{Cross-Model Generalization Analysis}

We evaluated several representative lightweight detectors on AIGIBench~\cite{li2025artificial}, which covers 25 diverse forgery categories ranging from GANs and DMs to artistic style transfer and social media image manipulations. 

\begin{table}[!t]
    \centering
    \caption{Detection accuracy (\%) of lightweight detectors on aigibench across different data categories.}
    \resizebox{\linewidth}{!}{
    \begin{tabular}{c|c|ccc|c|c}
        \toprule
        Model & Real & GAN-based & DM-based & In-the-Wild & Fake & Overall  \\ 
        \midrule
        UnivFD~\cite{ojha2023towards} & 95.03 & 66.18 & 14.60 & 23.99 & 28.53 & 61.81  \\ 
        PatchCraft~\cite{zhong2023rich} & 72.11 & 99.23 & 58.14 & 5.58 & 51.55 & 61.84  \\ 
        NPR~\cite{tan2024rethinking} & 91.19 & 94.99 & 81.12 & 7.53 & 62.45 & 76.84  \\ 
        FreqNet~\cite{tan2024frequency} & 85.03 & 69.00 & 50.89 & 14.57 & 44.11 & 64.59  \\ 
        DRCT~\cite{chen2024drct} & 38.74 & 83.88 & 90.18 & 89.00 & 88.47 & 63.58  \\ 
        AIDE~\cite{yan2024sanity} & 81.31 & 67.09 & 82.34 & 36.11 & 65.43 & 73.37  \\ 
        DiffAIGD~\cite{zhong2025beyond} & 47.02 & 98.41 & 99.04 & 67.73 & 89.68 & 68.33  \\ 
        Effort~\cite{yan2024effort} & 61.93 & 95.89 & 94.57 & 54.61 & 83.09 & 72.50  \\ 
        DDA~\cite{chen2025dual} & 94.45 & 79.13 & 72.03 & 37.28 & 63.33 & 78.90 \\ 
        D\textsuperscript{3}~\cite{yang2025d} & 86.94 & 88.67 & 84.50 & 38.72 & 71.92 & 79.44  \\ 
        \bottomrule
    \end{tabular}
    }
    \label{tab_per_split}
    \vspace{-1em}
\end{table}

Table~\ref{tab_per_split} summarizes the detection accuracy of several representative lightweight detectors across different generative categories.
All detectors are evaluated using their official pretrained models without additional fine-tuning on AIGIBench, ensuring a fair assessment.
As shown in Table~\ref{tab_per_split}, detectors trained on GAN-based data, such as UnivFD and PatchCraft, achieve relatively high accuracy on GAN images, but suffer a severe decline on DM-based images.
In contrast, diffusion-aware detectors such as DRCT and Effort perform well on all fake subsets, at the cost of a significantly lower accuracy on real images, reflecting over-sensitivity to anomaly. 
Moreover, most detectors exhibit consistently poor performance on in-the-wild fake images.
These results highlight the limitations of existing detectors regarding cross-model generalization.

\subsection{Prompt Engineering} \label{aigibench-dev}
To fully leverage the generalization potential of Multimodal Large Language Models for AI-generated image detection, we adopt a zero-shot setting in which MLLMs are prompted to perform detection.
Representative open-source MLLMs from the Qwen-VL series are considered.

As MLLMs rely solely on textual instructions in zero-shot scenarios, we design a structured prompt engineering framework.
Concretely, each prompt is composed of the system prompt, the question prompt, and the output prompt.

\begin{itemize}
    \item \textbf{System Prompts (6 options)}: 
    Inform the model of its role and background knowledge.
    Options include an empty prompt, a basic detection-expert role, a progressive inclusion of AIGC detection guidelines, and a comprehensive inference rule covering both image generation and editing-based manipulations. 
    \item \textbf{Question Prompts (7 options)}: 
    Inform the model of the task formulation.
    Options include covering binary classification questions, fine-grained semantic analysis, pixel-level artifact inspections, logical consistency checks, and balanced phrasing to mitigate output bias.
    \item \textbf{Output Prompts (4 options)}: 
    Regulate the response format returned by the model.
    Options include binary decisions, brief technical justifications, structured reasoning chains, and confidence-annotated outputs.
\end{itemize}

\begin{figure}[!t]
    \centering
    \includegraphics[width=\linewidth]{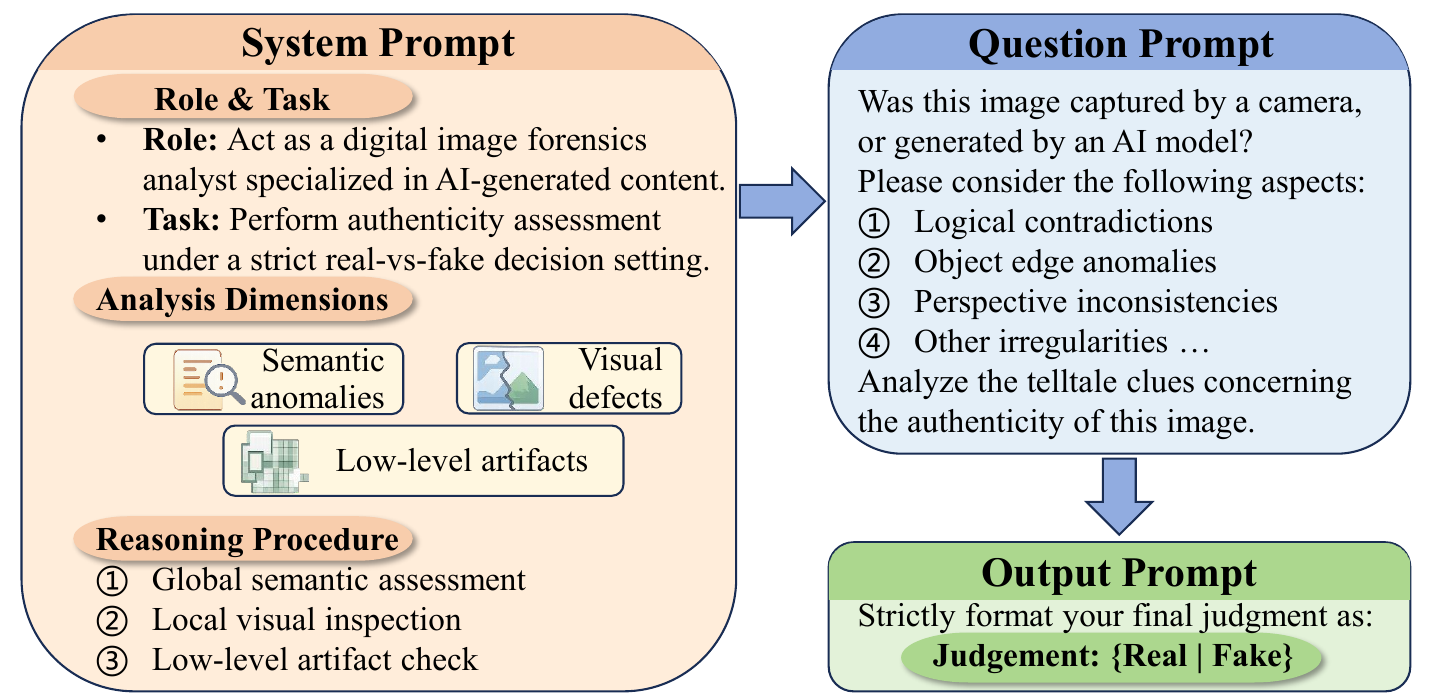}
    \caption{An example of the structured prompt used for MLLMs}
    \label{fig:prompt_structure}
\end{figure}

There are altogether $6\times7\times4=168$ prompts. 
We iterate over all prompts on a balanced subset constructed from AIGIBench, denoted as \textbf{AIGIBench-Dev}.
Specifically, this set is formed by randomly sampling 50 real images and 50 fake images from each of the 25 subsets in AIGIBench, resulting in 2,500 images in total.
The prompt configuration that achieves the highest detection accuracy on AIGIBench-Dev is then fixed and adopted for all subsequent experiments.
An illustrative example of the prompt is shown in Fig.~\ref{fig:prompt_structure}.

\subsection{Fuzzy Decision Tree}

MLLMs focus on high-level semantics, while lightweight detectors target low-level artifacts, making their detection results complementary. 
To exploit this, we adopt a fuzzy decision tree-based ensemble that adaptively integrates predictions from both types of models.
Each node of the decision tree represents a fuzzy predicate that involves several basic detectors (a basic detector is either an MLLM with an appropriate prompt or a lightweight detector). 
The generation of the fuzzy decision tree follows a greedy strategy guided by information gain. 
Given the current node, for each input image $x$, we collect prediction outputs from all $M$ models and unify them into a feature vector $\mathbf{f}(x) = [f_{1}(x), f_{2}(x), \cdots, f_{M}(x)]\in [0,1]^{M}$, where $f_{m}(x)$ is the likelihood that the image $x$ is fake judged by the $m$-th model. 
For consistency with continuous model outputs, binary predictions in $\left\{0,1\right\}$ are converted to soft labels, with $0.0$ for real and $1.0$ for fake.

Each node is featured by the following triplet: 
\begin{itemize}
    \item \textbf{A subset of detectors} $\mathcal{S} \subseteq \{f_{1}, \cdots, f_{M}\}$ is a subset of detectors that functions at the current node.
    \item \textbf{Ensemble operator} $\phi \in \{\texttt{mean}, \texttt{min}, \texttt{max}, \texttt{median}\}$ is the operator that combines the outputs of detectors in $\mathcal{S}$.
    \item \textbf{Threshold} $\tau \in [0,1]$ is the decision threshold at the current node, i.e., images whose prediction is larger than $\tau$ would be considered as forgery and passed to the next node, and \emph{vice versa}.
\end{itemize}
Without loss of generality, the root node can be considered as $(\emptyset,\texttt{max},0.5)$. 
For a node whose configuration is $(\mathcal{S}, \phi, \tau)$, the local prediction $\hat{p}(x)$ for sample $x$ is computed as:
\begin{equation}
\hat{p}(x) = \phi\left(\left[f(x) | f \in \mathcal{S}\right]\right),
\end{equation}
images whose local prediction is lower than $\tau$ are temporarily considered as real and are passed to the left (real) child node; the others are passed to the right (fake) child node. 

At each node, the information gain provided by a candidate configuration $(\mathcal{S}, \phi, \tau)$ is measured in the improvement in classification accuracy over the majority baseline:
\begin{equation}
\text{Gain}(\mathcal{S}, \phi, \tau) = \text{Accuracy}_{\text{split}} - \text{Accuracy}_{\text{majority}}
\end{equation}
where $\text{Accuracy}_{\text{majority}}$ denotes the accuracy achieved by returning the label of the dominant class within the current node and $\text{Accuracy}_{\text{split}}$ is the accuracy obtained by splitting samples according to the ensemble operator and threshold $\tau$.

During the growth of this fuzzy decision tree, a node would be split only if it introduces a non-trivial partition (i.e., the number of images belong to both child nodes are larger than a threshold) and splitting this node achieves a significant accuracy gain over the majority prediction baseline (when $\text{Gain}(\mathcal{S}, \phi, \tau) > 0$). 
The optimal split is exhaustively searched, and the fuzzy decision tree grows recursively until no more splits are available or all nodes meet a stopping criterion (e.g., the depth of the tree meets the maximum).

With the ensemble operator, the decision tree can process the outputs of multiple detectors in a fuzzy manner. 
For example, when $\mathcal{S}=\left\{f_{1},f_{2},f_{3}\right\}$ and the ensemble operator is $\phi=\texttt{max}$, the fuzzy logic predicate corresponding to the current node can be interpreted as:
\begin{equation}
\label{equation:rule}
\begin{aligned}
\textbf{IF } &f_{1} \text{ suggests that } x \text{ is fake} \textbf{ OR }f_{2} \text{ suggests that } x \text{ is fake} \\
&\textbf{ OR } f_{3} \text{ suggests that } x \text{ is fake} \textbf{ THEN } x \text{ is fake }, 
\end{aligned}
\end{equation}
in which all sentences are fuzzy logic. 
Using \texttt{min} operator as $\phi$ is tantamount to replacing \textbf{OR} with \textbf{AND} in Eq.~\eqref{equation:rule}. 

During inference, we first collect the predictions from all available detectors. 
Then we traverse the fuzzy decision tree from the root node by iteratively applying the fuzzy rule within each node. 
The input image would fall into a unique leaf node, which yields the final prediction. 
As every node in the tree is characterized by a model set, an ensemble operator, and a threshold, the inference is naturally interpretable.
\section{Experiments and discussions} \label{section:4}

\subsection{Settings}

\subsubsection{Datasets}

We evaluate the proposed detection framework on six benchmarks~\cite{li2025artificial, zhu2023genimage, zhong2023rich, cavia2024real, li2025fakebench, yan2024sanity}. 

The fuzzy decision tree and all comparative ensemble strategies are trained on the same small development set AIGIBench-Dev, ensuring a fair and controlled comparison across different fusion paradigms.

\subsubsection{Prompts for MLLMs}

\begin{figure}[!t]
    \centering
    \includegraphics[width=.9\linewidth]{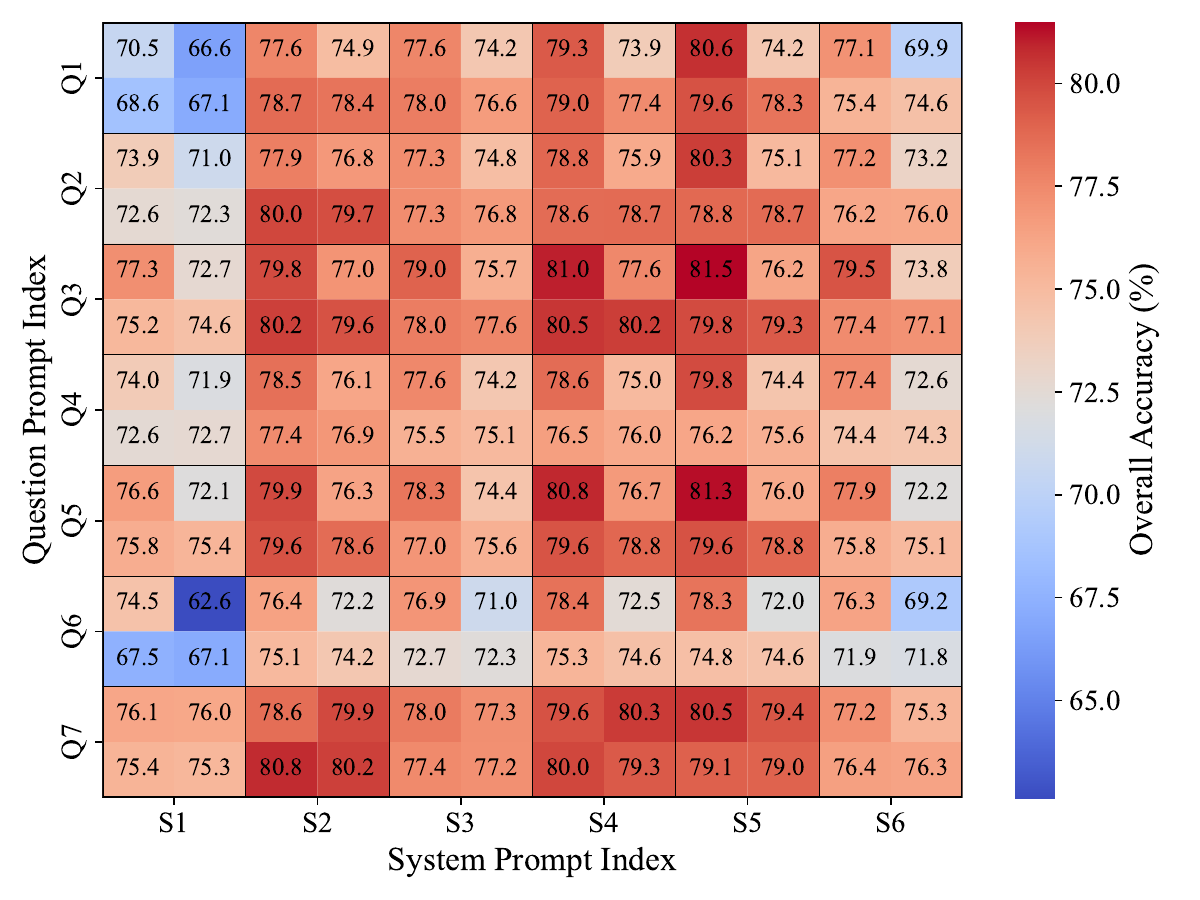}
    \caption{Accuracy of Qwen3-VL-32B-Instruct with different prompts. 
    Each cell is subdivided into four subcells, representing four output prompts.}
    \label{fig:prompt-heatmap}
\end{figure}

For each MLLM, we evaluated all 168 prompt combinations on AIGIBench-Dev and adopted the optimal configuration.
Fig.~\ref{fig:prompt-heatmap} illustrates the prompt-wise performance of Qwen3-VL-32B.
It is observed that the performance of MLLMs heavily relies on the prompt configuration. 
When the system and question prompts clearly articulate the task background and formulation, the model's discriminative performance significantly improves. 

Notably, most prompt variants still achieve reasonable accuracy, indicating that the model’s zero-shot capability is stable and does not depend on overly specific prompt configurations.

\subsubsection{Configurations of the fuzzy decision tree}

The key hyperparameters of the proposed fuzzy decision tree are summarized as follows.
Concretely, \textit{max\_split\_models} ($s$) determines the maximum number of base detectors participating in ensembling at each node, \textit{min\_samples} ($m$) determines the minimal number of images that a node might hold and prevents overfitting,  \textit{max\_depth} ($d$) limits the maximum depth of the tree, and \textit{thr\_grid\_size} ($g$) determines the granularity of the threshold search during node optimization.

For constructing the fuzzy decision tree, we select a set of representative detectors, including three lightweight models (NPR, DRCT, and AIDE) and three MLLMs (Qwen2-VL-7B, Qwen3-VL-8B, and Qwen3-VL-32B).
We train the fuzzy decision tree on AIGIBench-Dev and determine the hyperparameters based on classification accuracy; the final configuration is fixed as $s=3$, $m=0$, $d=4$, and $g=10$.

\subsection{Performance Comparison}

\begin{table*}[!t]
    \centering
    \caption{Comparison of baselines and the fuzzy decision tree. Here, $\dagger$ denotes models used as base components for the ensemble strategies.
    \textbf{Bold} and \underline{underline} indicate the best and second-best performance in each column, respectively. 
    \textbf{Overall} denotes the sample-weighted accuracy, while \textbf{avg.} and \textbf{std.} are computed across benchmarks.
    \textbf{Robustness} reports the mean accuracy under blur/jpeg/resize on aigibench-dev.
    \textbf{aigibench} is evaluated on the remaining test set excluding aigibench-dev.
    }
    \label{tab:baseline}
    \begin{tabular}{l|cccccc|ccc|c}
        \toprule
        \textbf{Model} 
        & \textbf{AIGIBench} 
        & \textbf{GenImage} 
        & \textbf{AIGCDet} 
        & \textbf{WildRF} 
        & \textbf{FakeBench} 
        & \textbf{Chameleon} 
        & \textbf{Overall}
        & \textbf{Avg.} 
        & \textbf{Std.}
        & \textbf{Robustness} \\
        \midrule
        
        \multicolumn{11}{l}{\textbf{Lightweight Detectors}} \\
        NPR$^\dagger$~\cite{tan2024rethinking} & 76.84 & 93.71 & 93.90 & 67.52 & 61.37 & 59.25 & 84.27 & 75.43 & 14.14 & 71.35 \\
        DRCT$^\dagger$~\cite{chen2024drct}      & 63.58 & 89.69 & 84.08 & 53.46 & 85.47 & 53.86 & 74.77 & 71.69 & 15.18 & 65.89 \\
        AIDE$^\dagger$~\cite{yan2024sanity}     & 73.37 & 87.40 & 83.07 & 67.28 & 65.37 & 62.61 & 78.45 & 73.18 & 9.20 & 70.41 \\
        UnivFD~\cite{ojha2023towards}      & 61.81 & 69.71 & 74.94 & 57.61 & 71.65 & 57.40 & 67.27 & 65.52 & 6.91 & 57.90 \\
        PatchCraft~\cite{zhong2023rich}    & 61.84 & 67.49 & 77.92 & 65.16 & 50.37 & 58.08 & 67.56 & 63.48 & 8.48 & 66.68 \\
        FreqNet~\cite{tan2024frequency}    & 64.59 & 73.84 & 78.60 & 59.09 & 58.95 & 58.78 & 70.32 & 65.64 & 7.86 & 66.75 \\
        DiffAIGD~\cite{zhong2025beyond}    & 68.33 & 95.87 & 80.45 & 75.31 & \textbf{88.15} & 59.83 & 77.37 & 77.99 & 11.97 & 83.03 \\
        Effort~\cite{yan2024effort}        & 72.50 & 93.95 & 89.24 & 67.24 & 71.15 & 52.42 & 80.81 & 74.42 & 13.85 & 73.51 \\
        DDA~\cite{chen2025dual}            & 78.90 & 88.47 & 88.29 & 84.74 & 84.77 & \textbf{84.82} & 84.09 & 85.00 & \textbf{3.17} & 80.57 \\
        D$^3$~\cite{yang2025d}             & 79.44 & 97.08 & \textbf{96.24} & 63.27 & 85.17 & 65.93 & 87.38 & 81.19 & 13.24 & 80.76 \\
        \midrule
        \multicolumn{11}{l}{\textbf{Multimodal Large Language Models (Zero-Shot)}} \\
        Qwen2-VL-7B$^\dagger$    & 72.99 & 70.16 & 66.58 & 91.57 & 80.97 & 76.93 & 70.86 & 76.53 & 8.15 & 81.52 \\
        Qwen2.5-VL-7B & 64.91 & 56.58 & 57.97 & 92.09 & 75.58 & 66.80 & 61.49 & 68.99 & 12.07 & 64.30 \\
        Qwen2.5-VL-32B                  & 74.71 & 69.02 & 68.19 & 95.21 & 79.85 & 74.48 & 71.73 & 76.91 & 9.06 & 75.53 \\
        Qwen3-VL-8B$^\dagger$    & 76.05 & 75.53 & 71.75 & \textbf{95.81} & 85.43 & 81.44 & 75.13 & 81.00 & 7.95 & 76.04 \\
        Qwen3-VL-32B$^\dagger$                    & 77.18 & 75.69 & 71.88 & \underline{95.65} & 82.97 & 81.05 & 75.63 & 80.74 & 7.57 & 79.96 \\
        
        \midrule
        \multicolumn{11}{l}{\textbf{Ensemble Strategies}} \\
        Voting                 & 79.47 & 95.55 & 91.99 & 91.73 & 87.13 & 79.99 & 86.69 & 87.64 & 6.11 & 85.43 \\
        Logistic Regression  & 83.37 & 96.80 & 94.53 & 92.89 & 85.87 & 79.27 & 89.33 & 88.79 & 6.36 & 85.96 \\
        Linear SVM           & 83.53 & 96.75 & 94.85 & 92.69 & 85.75 & 79.80 & 89.51 & 88.90 & 6.23 & 86.01 \\
        LightGBM             & \underline{83.95} & 96.38 & 95.10 & 92.01 & 84.55 & 79.58 & \underline{89.66} & 88.60 & 6.24 & 86.51 \\
        Bagging Tree         & 83.60 & 97.04 & 94.35 & 93.21 & 84.58 & 80.48 & 89.47 & 88.88 & 6.22 & 86.84 \\
        Random Forest        & 83.64 & \underline{97.10} & 94.77 & 92.77 & 84.60 & 80.61 & 89.63 & \underline{88.92} & 6.21 & \underline{86.92} \\
        \midrule
        \multicolumn{11}{l}{\textbf{Fuzzy Decision Tree (Ours)}} \\
        Ours & \textbf{84.04} & \textbf{97.12} & \underline{95.49} & 93.87 & \underline{87.58} & \underline{82.12} & \textbf{90.14} & \textbf{90.04} & \underline{5.76} & \textbf{87.40} \\
        \bottomrule
        \end{tabular}
\end{table*}

\begin{table}[!t]
    \centering
    \caption{
    Ablation study of the proposed fuzzy decision tree.
    }
    \label{tab:ablation_study}
    \begin{tabular}{cc|cccc}
        \toprule
        \textbf{Lightweight} & \textbf{MLLMs}
        & \textbf{Overall}
        & \textbf{Avg.}
        & \textbf{Std.}
        & \textbf{Robustness} \\
        \midrule
        \checkmark &            & 87.32 & 81.11 & 11.75 & 79.99 \\
                   & \checkmark & 77.83 & 82.08 & 6.22  & 84.69 \\
        \checkmark & \checkmark & \textbf{90.14} & \textbf{90.04} & \textbf{5.76} & \textbf{87.40} \\
        \bottomrule
    \end{tabular}
    \vspace{-1em}
\end{table}

\begin{figure*}[!t]
    \centering
    \includegraphics[width=.95\linewidth]{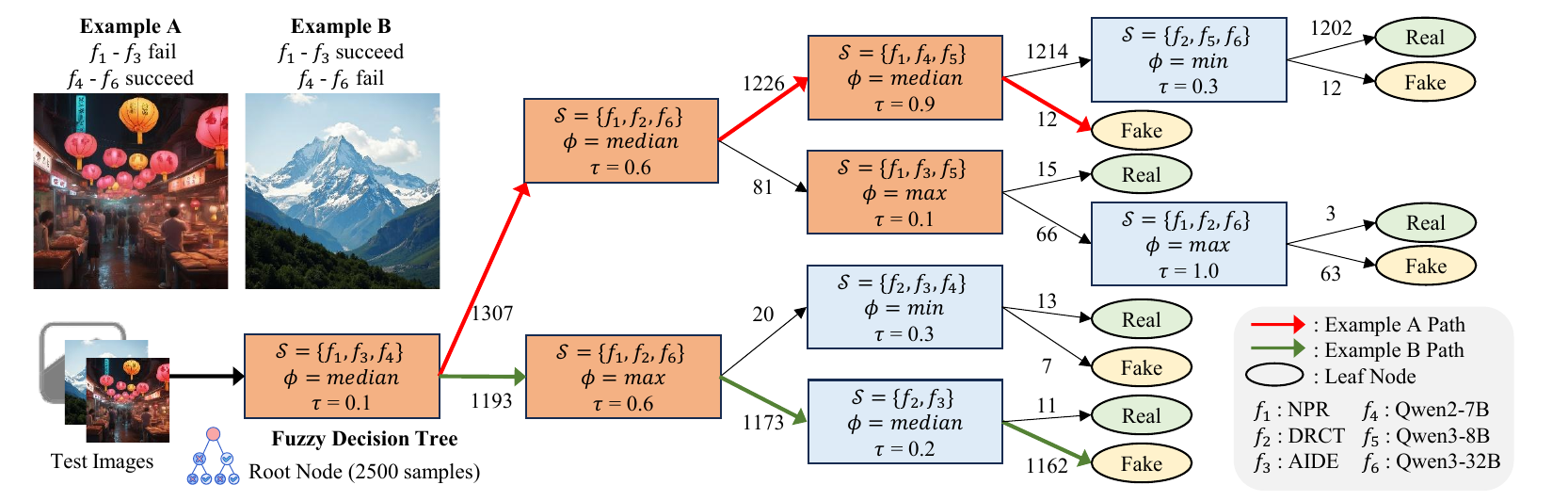}
    \caption{Visualized structure of the fuzzy decision tree with inference paths for two representative fake image examples. Each internal node is parameterized by a fuzzy predicate $(\mathcal{S}, \phi, \tau)$, where a selected subset of detectors $\mathcal{S}$ is fused via an ensemble operator $\phi$ and compared against a threshold $\tau$.}
    \label{fig:fuzzy_tree}
\end{figure*}

We considered three categories of baselines for comparison: lightweight detectors, zero-shot MLLMs, and six vanilla ensemble strategies. 
Table~\ref{tab:baseline} summarizes their detection accuracy across six benchmarks, together with overall performance.

\subsubsection{Baselines}

\textbf{Baseline I: Lightweight detectors.} 
We investigated ten representative lightweight AI-generated image detectors, including recent state-of-the-art methods~\cite{zhong2025beyond, yan2024effort, chen2025dual, yang2025d}.
Among them, NPR, DRCT, and AIDE are additionally adopted as base components of the proposed fuzzy decision tree, while the remaining detectors serve as competitive standalone baselines.
All detectors were instantiated using the official pretrained weights without any fine-tuning or adaptation. 

As shown in Table~\ref{tab:baseline}, the performance of lightweight detectors exhibited substantial variations across benchmarks, suggesting that these detectors tend to overfit to specific generative patterns seen during training and struggle to generalize across diverse synthesis techniques.

\textbf{Baseline II: MLLMs.}  
Table~\ref{tab:baseline} demonstrates that MLLMs could achieve a high average accuracy with lower performance variance compared with lightweight detectors.
Although the prompt engineering was conducted solely on AIGIBench-Dev, the performance remained stable across the other benchmarks.
Therefore, systematic prompt engineering for MLLMs can boost generalizable detection capability.

Notably, MLLMs consistently outperform most lightweight detectors on in-the-wild datasets such as WildRF and Chameleon.
In contrast, their performance on artifact-sensitive benchmarks such as GenImage is generally inferior.
These observations indicate that MLLMs are more effective at capturing high-level semantic inconsistencies, but lack sufficient sensitivity to subtle low-level generation artifacts.

\textbf{Baseline III: Ensembles.}
We considered six vanilla ensemble strategies, including majority voting, linear models, and tree-based methods.
All ensemble methods were trained on AIGIBench-Dev using identical input features to ensure a fair comparison.
For each strategy, the optimal performance is reported in Table~\ref{tab:baseline}.

Despite their simplicity, ensemble methods significantly outperformed single-model baselines regarding both accuracy and stability. 
This performance gain demonstrates that the detection results from MLLMs are highly complementary to those from lightweight artifact-aware detectors.

\subsubsection{Comparisons}

As shown in Table~\ref{tab:baseline}, the proposed fuzzy decision tree achieves the state-of-the-art overall performance across benchmarks.
Compared with vanilla ensemble baselines, our method can adaptively learn the optimal combination of detectors and fusion rules at each node, resulting in more flexible and fine-grained decision boundaries.
These results demonstrate the superior effectiveness and generalization capability of the proposed framework for general AI-generated content detection.

\subsection{Robustness Analysis} \label{section:robust}

We evaluate robustness under common post-processing operations by applying Gaussian blur, JPEG compression, and image resizing, each with multiple severity levels, on AIGIBench-Dev.
Table~\ref{tab:baseline} reports the average accuracy under post-processing perturbations.

Lightweight detectors suffer noticeable performance degradation under post-processing perturbations, reflecting their sensitivity to low-level image statistics, whereas MLLMs exhibit improved robustness, likely due to their reliance on high-level semantic reasoning.
Notably, the proposed fuzzy decision tree achieves the highest average robustness among all compared methods under realistic image distortions.

\subsection{Ablation Study} \label{section:ablation}

Table~\ref{tab:ablation_study} presents an ablation study of the proposed fuzzy decision tree by selectively enabling lightweight detectors and MLLMs.
Using only lightweight detectors yields relatively high overall accuracy but suffers from large variance and limited robustness, while using only MLLMs improves robustness and stability at the cost of overall accuracy.
When both components are enabled, the full fuzzy decision tree consistently achieves the best performance across all metrics.

\subsection{Interpretability and Case Studies}

Fig.~\ref{fig:fuzzy_tree} visualizes the learned fuzzy decision tree and the inference paths for two representative fake images.
For Example~A, most lightweight detectors fail to detect the forgery, whereas MLLMs correctly identify semantic anomalies.
Conversely, Example~B exhibits strong semantic coherence that misleads MLLMs, while lightweight detectors capture subtle low-level texture artifacts.
In both cases, the fuzzy decision tree correctly classifies the images as fake by effectively integrating the complementary strengths of different types of detectors in an interpretable decision-making process.

\section{Conclusion} \label{section:5}

In this paper, we propose a generalizable and interpretable framework for AI-generated image detection by synergistically integrating lightweight detectors and MLLMs through a fuzzy decision tree. 
This design enables adaptive fusion of low-level artifact cues and high-level semantic reasoning, effectively addressing the limitations of existing methods that either overfit to model-specific artifacts or lack fine-grained perceptual sensitivity.
Extensive experiments on multiple benchmarks demonstrate that our method achieves the state-of-the-art overall performance, strong cross-model generalization, and enhanced robustness.
Moreover, the fuzzy decision tree provides an interpretable decision-making process.
Our work provides a promising path toward building robust, generalizable, and extensible solutions for real-world AI-generated image detection.

\bibliographystyle{IEEEbib}
\bibliography{icme2026references}

@inproceedings{ojha2023towards,
  title={Towards universal fake image detectors that generalize across generative models},
  author={Ojha, Utkarsh and Li, Yuheng and others},
  booktitle={CVPR},
  pages={24480--24489},
  year={2023}
}

@article{zhong2023rich,
  title={Rich and poor texture contrast: A simple yet effective approach for ai-generated image detection},
  author={Zhong, Nan and Xu, Yiran and others},
  journal={CoRR},
  year={2023}
}

@inproceedings{tan2024rethinking,
  title={Rethinking the up-sampling operations in cnn-based generative network for generalizable deepfake detection},
  author={Tan, Chuangchuang and Zhao, Yao and others},
  booktitle={CVPR},
  pages={28130--28139},
  year={2024}
}

@inproceedings{tan2024frequency,
  title={Frequency-aware deepfake detection: Improving generalizability through frequency space domain learning},
  author={Tan, Chuangchuang and Zhao, Yao and others},
  booktitle={AAAI},
  pages={5052--5060},
  year={2024}
}

@inproceedings{chen2024drct,
  title={Drct: Diffusion reconstruction contrastive training towards universal detection of diffusion generated images},
  author={Chen, Baoying and Zeng, Jishen and others},
  booktitle={ICML},
  year={2024}
}

@inproceedings{zhong2025beyond,
  title={Beyond Generation: A Diffusion-based Low-level Feature Extractor for Detecting AI-generated Images},
  author={Zhong, Nan and Chen, Haoyu and others},
  booktitle={CVPR},
  pages={8258--8268},
  year={2025}
}

@article{yan2024sanity,
  title={A sanity check for ai-generated image detection},
  author={Yan, Shilin and Li, Ouxiang and others},
  journal={arXiv preprint arXiv:2406.19435},
  year={2024}
}

@article{yan2024effort,
  title={Effort: Efficient Orthogonal Modeling for Generalizable AI-Generated Image Detection},
  author={Yan, Zhiyuan and Wang, Jiangming and others},
  journal={arXiv preprint arXiv:2411.15633},
  year={2024}
}

@inproceedings{yang2025d,
  title={D\^{} 3: Scaling Up Deepfake Detection by Learning from Discrepancy},
  author={Yang, Yongqi and Qian, Zhihao and others},
  booktitle={CVPR},
  pages={23850--23859},
  year={2025}
}

@article{chen2025dual,
  title={Dual Data Alignment Makes AI-Generated Image Detector Easier Generalizable},
  author={Chen, Ruoxin and Xi, Junwei and others},
  journal={arXiv preprint arXiv:2505.14359},
  year={2025}
}

@article{li2025artificial,
  title={Is Artificial Intelligence Generated Image Detection a Solved Problem?},
  author={Li, Ziqiang and Yan, Jiazhen and others},
  journal={arXiv preprint arXiv:2505.12335},
  year={2025}
}

@article{zhu2023genimage,
  title={Genimage: A million-scale benchmark for detecting ai-generated image},
  author={Zhu, Mingjian and Chen, Hanting and others},
  journal={Advances in Neural Information Processing Systems},
  volume={36},
  pages={77771--77782},
  year={2023}
}

@article{li2025fakebench,
  title={Fakebench: Probing explainable fake image detection via large multimodal models},
  author={Li, Yixuan and Liu, Xuelin and others},
  journal={IEEE Transactions on Information Forensics and Security},
  year={2025},
  publisher={IEEE}
}

@article{cavia2024real,
  title={Real-time deepfake detection in the real-world},
  author={Cavia, Bar and Horwitz, Eliahu and others},
  journal={arXiv preprint arXiv:2406.09398},
  year={2024}
}

@article{ho2020denoising,
  title={Denoising diffusion probabilistic models},
  author={Ho, Jonathan and Jain, Ajay and others},
  journal={Advances in neural information processing systems},
  volume={33},
  pages={6840--6851},
  year={2020}
}

@article{goodfellow2020generative,
  title={Generative adversarial networks},
  author={Goodfellow, Ian and Pouget-Abadie, Jean and others},
  journal={Communications of the ACM},
  volume={63},
  number={11},
  pages={139--144},
  year={2020},
  publisher={ACM New York, NY, USA}
}

@article{tan2025veritas,
  title={Veritas: Generalizable Deepfake Detection via Pattern-Aware Reasoning},
  author={Tan, Hao and Lan, Jun and others},
  journal={arXiv preprint arXiv:2508.21048},
  year={2025}
}

@article{huang2024ffaa,
  title={Ffaa: Multimodal large language model based explainable open-world face forgery analysis assistant},
  author={Huang, Zhengchao and Xia, Bin and others},
  journal={arXiv preprint arXiv:2408.10072},
  year={2024}
}

@article{xu2024fakeshield,
  title={Fakeshield: Explainable image forgery detection and localization via multi-modal large language models},
  author={Xu, Zhipei and Zhang, Xuanyu and others},
  journal={arXiv preprint arXiv:2410.02761},
  year={2024}
}

@inproceedings{HuangHLH00W0C25,
  author       = {Zhenglin Huang and Jinwei Hu and others},
  title        = {{SIDA:} Social Media Image Deepfake Detection, Localization and Explanation
                  with Large Multimodal Model},
  booktitle    = {CVPR},
  year         = {2025},
}

@misc{cnn2024deepfake,
  author       = {Kathleen Magramo},
  title        = {Deepfake scam tricks Hong Kong firm into paying out \$25 million},
  howpublished = {\url{https://edition.cnn.com/2024/02/04/asia/deepfake-cfo-scam-hong-kong-intl-hnk}},
  year         = {2024},
  note         = {Accessed: 2025-06-30}
}

@inproceedings{liu2024forgery,
  title={Forgery-aware adaptive transformer for generalizable synthetic image detection},
  author={Liu, Huan and Tan, Zichang and others},
  booktitle={CVPR},
  pages={10770--10780},
  year={2024}
}

@article{han2023three,
  title={A three-way classification with fuzzy decision trees},
  author={Han, Xiaoyu and Zhu, Xiubin and others},
  journal={Applied Soft Computing},
  volume={132},
  pages={109788},
  year={2023},
  publisher={Elsevier}
}

@article{zhu2024development,
  title={A development of fuzzy-rule-based regression models through using decision trees},
  author={Zhu, Xiubin and Hu, Xingchen and others},
  journal={IEEE Transactions on Fuzzy Systems},
  volume={32},
  number={5},
  pages={2976--2986},
  year={2024},
  publisher={IEEE}
}

\end{document}